\documentclass[10pt]{article}
\usepackage{acl2001,times}
\title{Organizing Encyclopedic Knowledge based on the Web and its
Application to Question Answering}

\author{Atsushi Fujii \\
  University of Library and \\ Information Science \\
  1-2 Kasuga, Tsukuba \\
  305-8550, Japan \\
  CREST, Japan Science and \\ Technology Corporation \\
  {\tt fujii@ulis.ac.jp} \And
  Tetsuya Ishikawa \\
  University of Library and \\ Information Science \\
  1-2 Kasuga, Tsukuba \\
  305-8550, Japan \\
  {\tt ishikawa@ulis.ac.jp}
}

\date{}

\newcommand{\etal}{et~al.}

\newcommand{\eq}[1]{(\ref{#1})}

\input{psfig.tex}

\begin{document}
\maketitle
\begin{abstract}
  We propose a method to generate large-scale encyclopedic knowledge,
  which is valuable for much NLP research, based on the Web. We first
  search the Web for pages containing a term in question.  Then we use
  linguistic patterns and HTML structures to extract text fragments
  describing the term. Finally, we organize extracted term
  descriptions based on word senses and domains. In addition, we apply
  an automatically generated encyclopedia to a question answering
  system targeting the Japanese Information-Technology Engineers
  Examination.
\end{abstract}

\section{Introduction}
\label{sec:introduction}

Reflecting the growth in utilization of the World Wide Web, a number
of Web-based language processing methods have been proposed within the
natural language processing (NLP), information retrieval (IR) and
artificial intelligence (AI) communities. A sample of these includes
methods to {\em extract\/} linguistic
resources~\cite{fujii:acl-2000,resnik:acl-99,soderland:kdd-97}, {\em
retrieve\/} useful information in response to user
queries~\cite{etzioni:ai-magazine-97,mccallum:ijcai-99} and {\em
mine/discover\/} knowledge latent in the Web~\cite{inokuchi:pakdd-99}.

In this paper, mainly from an NLP point of view, we explore a method
to produce linguistic resources. Specifically, we enhance the method
proposed by Fujii and Ishikawa~\shortcite{fujii:acl-2000}, which
extracts encyclopedic knowledge (i.e., term descriptions) from the
Web.

In brief, their method searches the Web for pages containing a term in
question, and uses linguistic expressions and HTML layouts to extract
fragments describing the term. They also use a language model to
discard non-linguistic fragments.  In addition, a clustering method is
used to divide descriptions into a specific number of groups.

On the one hand, their method is expected to enhance existing
encyclopedias, where vocabulary size is relatively limited, and
therefore the {\em quantity\/} problems has been resolved.

On the other hand, encyclopedias extracted from the Web are not
comparable with existing ones in terms of {\em quality}.  In
hand-crafted encyclopedias, term descriptions are carefully organized
based on domains and word senses, which are especially effective for
human usage.  However, the output of Fujii's method is simply a set of
unorganized term descriptions.  Although clustering is optionally
performed, resultant clusters are not necessarily related to explicit
criteria, such as word senses and domains.

To sum up, our belief is that by combining {\em extraction\/} and {\em
organization\/} methods, we can enhance both quantity and quality of
Web-based encyclopedias.

Motivated by this background, we introduce an organization model to
Fujii's method and reformalize the whole framework.  In other words,
our proposed method is not only extraction but {\em generation\/} of
encyclopedic knowledge.

Section~\ref{sec:system_design} explains the overall design of our
encyclopedia generation system, and Section~\ref{sec:organization}
elaborates on our organization model.  Section~\ref{sec:application}
then explores a method for applying our resultant encyclopedia to NLP
research, specifically, question answering.
Section~\ref{sec:experimentation} performs a number of experiments to
evaluate our methods.

\section{System Design}
\label{sec:system_design}

\subsection{Overview}
\label{subsec:system_overview}

\begin{figure}[htbp]
  \begin{center}
    \leavevmode
    \psfig{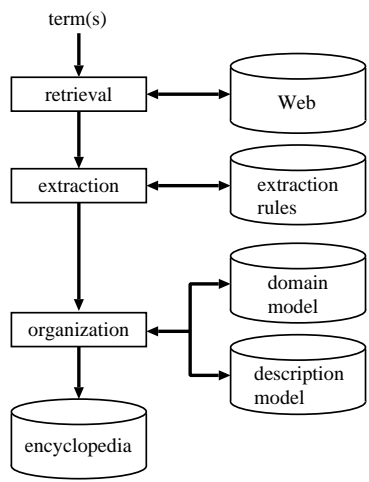}
  \end{center}
  \caption{The overall design of our Web-based encyclopedia generation
    system.}
  \label{fig:system}
\end{figure}

Figure~\ref{fig:system} depicts the overall design of our system,
which generates an encyclopedia for input terms.

Our system, which is currently implemented for Japanese, consists of
three modules: ``retrieval,'' ``extraction'' and ``organization,''
among which the organization module is newly introduced in this paper.
In principle, the remaining two modules (``retrieval'' and
``extraction'') are the same as proposed by Fujii and
Ishikawa~\shortcite{fujii:acl-2000}.

In Figure~\ref{fig:system}, terms can be submitted either on-line or
off-line. A reasonable method is that while the system periodically
updates the encyclopedia off-line, terms unindexed in the encyclopedia
are dynamically processed in real-time usage.  In either case, our
system processes input terms one by one.

We briefly explain each module in the following three sections,
respectively.

\subsection{Retrieval}
\label{subsec:retrieval}

The retrieval module searches the Web for pages containing an input
term, for which existing Web search engines can be used, and those
with broad coverage are desirable.

However, search engines performing query expansion are not always
desirable, because they usually retrieve a number of pages which do
not contain an input keyword.  Since the extraction module (see
Section~\ref{subsec:extraction}) analyzes the usage of the input term
in retrieved pages, pages not containing the term are of no use for our
purpose.

Thus, we use as the retrieval module ``Google,'' which is one of the
major search engines and does not conduct query
expansion\footnote{http://www.google.com/}.

\subsection{Extraction}
\label{subsec:extraction}

In the extraction module, given Web pages containing an input term,
newline codes, redundant white spaces and HTML tags that are not used
in the following processes are discarded to standardize the page
format.

Second, we approximately identify a region describing the term in the
page, for which two rules are used.

The first rule is based on Japanese linguistic patterns typically used
for term descriptions, such as ``X {\it toha\/} Y {\it dearu\/} (X is
Y).''  Following the method proposed by Fujii and
Ishikawa~\shortcite{fujii:acl-2000}, we semi-automatically produced 20
patterns based on the Japanese CD-ROM World
Encyclopedia~\cite{heibonsha:98}, which includes approximately 80,000
entries related to various fields.  It is expected that a region
including the sentence that matched with one of those patterns can be
a term description.

The second rule is based on HTML layout. In a typical case, a term in
question is highlighted as a heading with tags such as \verb|<DT>|,
\verb|<B>| and \verb|<Hx>| (``\verb|x|'' denotes a digit), followed by
its description. In some cases, terms are marked with the anchor
\verb|<A>| tag, providing hyperlinks to pages where they are
described.

Finally, based on the region briefly identified by the above method,
we extract a page fragment as a term description. Since term
descriptions usually consist of a logical segment (such as a
paragraph) rather than a single sentence, we extract a fragment that
matched with one of the following patterns, which are sorted according
to preference in descending order:
\begin{enumerate}
\item description tagged with \verb|<DD>| in the case where the term
  is tagged with \verb|<DT>|\footnote{\texttt{<DT>} and \texttt{<DD>} are
  inherently provided to describe terms in HTML.},
\item paragraph tagged with \verb|<P>|,
\item itemization tagged with \verb|<UL>|,
\item $N$ sentences, where we empirically set \mbox{$N = 3$}.
\end{enumerate}

\subsection{Organization}
\label{subsec:organization}

As discussed in Section~\ref{sec:introduction}, organizing information
extracted from the Web is crucial in our framework.  For this purpose,
we classify extracted term descriptions based on word senses and
domains.

Although a number of methods have been proposed to generate word
senses (for example, one based on the vector space
model~\cite{schutze:cl-98}), it is still difficult to accurately
identify word senses without explicit dictionaries that define sense
candidates.

In addition, since word senses are often associated with
domains~\cite{yarowsky:acl-95}, word senses can be consequently
distinguished by way of determining the domain of each description.
For example, different senses for ``pipeline (processing
method/transportation pipe)'' are associated with the computer and
construction domains (fields), respectively.

To sum up, the organization module classifies term descriptions based
on domains, for which we use domain and description models.  In
Section~\ref{sec:organization}, we elaborate on our organization
model.

\section{Statistical Organization Model}
\label{sec:organization}

\subsection{Overview}
\label{subsec:organization_overview}

Given one or more (in most cases more than one) descriptions for a
single input term, the organization module selects appropriate
description(s) for each domain related to the term.

We do not need all the extracted descriptions as final outputs,
because they are usually similar to one another, and thus are
redundant.

For the moment, we assume that we know {\it a priori\/} which domains
are related to the input term.

From the viewpoint of probability theory, our task here is to select
descriptions with greater probability for given domains.  The
probability for description $d$ given domain $c$, \mbox{$P(d|c)$}, is
commonly transformed as in Equation~\eq{eq:organization}, through
use of the Bayesian theorem.
\begin{equation}
  \label{eq:organization}
  P(d|c) = \frac{\textstyle P(c|d)\cdot P(d)}{\textstyle P(c)}
\end{equation}
In practice, $P(c)$ can be omitted because this factor is a constant,
and thus does not affect the relative probability for different
descriptions.

In Equation~\eq{eq:organization}, $P(c|d)$ models a probability that
$d$ corresponds to domain $c$. $P(d)$ models a probability that $d$
can be a description for the term in question, disregarding the
domain. We shall call them domain and description models, respectively.

To sum up, in principle we select $d$'s that are strongly associated
with a specific domain, and are likely to be descriptions themselves.

Extracted descriptions are not linguistically understandable in the
case where the extraction process is unsuccessful and retrieved pages
inherently contain non-linguistic information (such as special
characters and e-mail addresses).

To resolve this problem, Fujii and Ishikawa~\shortcite{fujii:acl-2000}
used a language model to filter out descriptions with low
perplexity. However, in this paper we integrated a description model,
which is practically the same as a language model, with an
organization model. The new framework is more understandable with
respect to probability theory.

In practice, we first use Equation~\eq{eq:organization} to compute
$P(d|c)$ for all the $c$'s predefined in the domain model. Then we
discard such $c$'s whose $P(d|c)$ is below a specific threshold.  As a
result, for the input term, related domains and descriptions are
simultaneously selected. Thus, we do not have to know {\it a priori\/}
which domains are related to each term.

In the following two sections, we explain methods to realize the
domain and description models, respectively.

\subsection{Domain Model}
\label{subsec:domain_model}

The domain model quantifies the extent to which description $d$ is
associated with domain $c$, which is fundamentally a categorization
task.  Among a number of existing categorization methods, we
experimentally used one proposed by Iwayama and
Tokunaga~\shortcite{iwayama:anlp-94}, which formulates $P(c|d)$ as in
Equation~(\ref{eq:domain_model}).
\begin{equation}
  \label{eq:domain_model}
  P(c|d) = P(c)\cdot\sum_{t}\frac{\textstyle P(t|c)\cdot
  P(t|d)}{\textstyle P(t)}
\end{equation}
Here, $P(t|d)$, $P(t|c)$ and $P(t)$ denote probabilities that word $t$
appears in $d$, $c$ and all the domains, respectively. We regard
$P(c)$ as a constant. While $P(t|d)$ is simply a relative frequency of
$t$ in $d$, we need predefined domains to compute $P(t|c)$ and $P(t)$.
For this purpose, the use of large-scale corpora annotated with
domains is desirable.

However, since those resources are prohibitively expensive, we used
the ``Nova'' dictionary for Japanese/English machine translation
systems\footnote{Produced by NOVA, Inc.}, which includes approximately
one million entries related to 19 technical fields as listed below:
\begin{quote}
  aeronautics,
  biotechnology,
  business,
  chemistry,
  computers,
  construction,
  defense,
  ecology,
  electricity,
  energy,
  finance,
  law,
  mathematics,
  mechanics,
  medicine,
  metals,
  oceanography,
  plants,
  trade.
\end{quote}

We extracted words from dictionary entries to estimate $P(t|c)$ and
$P(t)$, which are relative frequencies of $t$ in $c$ and all the
domains, respectively.  We used the ChaSen morphological
analyzer~\cite{matsumoto:chasen-97} to extract words from Japanese
entries.  We also used English entries because Japanese descriptions
often contain English words.

It may be argued that statistics extracted from dictionaries are
unreliable, because word frequencies in real word usage are missing.
However, words that are representative for a domain tend to be
frequently used in compound word entries associated with the domain,
and thus our method is a practical approximation.

\subsection{Description Model}
\label{subsec:desc_model}

The description model quantifies the extent to which a given page
fragment is feasible as a description for the input term.  In
principle, we decompose the description model into language and
quality properties, as shown in Equation~(\ref{eq:desc_model}).
\begin{equation}
  \label{eq:desc_model}
  P(d) = P_{L}(d)\cdot P_{Q}(d)
\end{equation}
Here, $P_{L}(d)$ and $P_{Q}(d)$ denote language and quality models,
respectively.

It is expected that the quality model discards incorrect or misleading
information contained in Web pages. For this purpose, a number of
quality rating methods for Web
pages~\cite{amento:sigir-2000,zhu:sigir-2000} can be used.

However, since Google (i.e., the search engine used in our system)
rates the quality of pages based on hyperlink information, and
selectively retrieves those with higher quality
\cite{brin:compnet-1998}, we tentatively regarded $P_{Q}(d)$ as a
constant. Thus, in practice the description model is approximated
solely with the language model as in Equation~(\ref{eq:lang_model}).
\begin{equation}
  \label{eq:lang_model}
  P(d) \approx P_{L}(d)
\end{equation}

Statistical approaches to language modeling have been used in much NLP
research, such as machine translation~\cite{brown:cl-93} and speech
recognition~\cite{bahl:ieee-tpami-1983}. Our model is almost the same
as existing models, but is different in two respects.

First, while general language models quantify the extent to which a
given word sequence is linguistically acceptable, our model also
quantifies the extent to which the input is acceptable as a term
description.  Thus, we trained the model based on an existing machine
readable encyclopedia.

We used the ChaSen morphological analyzer to segment the Japanese
CD-ROM World Encyclopedia~\cite{heibonsha:98} into words (we replaced
headwords with a common symbol), and then used the CMU-Cambridge
toolkit~\cite{clarkson:eurospeech-97} to model a word-based trigram.

Consequently, descriptions in which word sequences are more similar to
those in the World Encyclopedia are assigned greater probability
scores through our language model.

Second, $P(d)$, which is a product of probabilities for $N$-grams in
$d$, is quite sensitive to the length of $d$. In the cases of machine
translation and speech recognition, this problem is less crucial
because multiple candidates compared based on the language model are
almost equivalent in terms of length.

However, since in our case length of descriptions are significantly
different, shorter descriptions are more likely to be selected,
regardless of the quality.  To avoid this problem, we normalize $P(d)$
by the number of words contained in $d$.

\section{Application}
\label{sec:application}

\subsection{Overview}
\label{subsec:application_overview}

Encyclopedias generated through our Web-based method can be used in a
number of applications, including human usage, thesaurus
production~\cite{hearst:coling-92,nakamura:coling-88} and natural
language understanding in general.

Among the above applications, natural language understanding (NLU) is
the most challenging from a scientific point of view.  Current
practical NLU research includes dialogue, information extraction and
question answering, among which we focus solely on question answering
(QA) in this paper.

A straightforward application is to answer interrogative questions
like ``What is X?'' in which a QA system searches the encyclopedia
database for one or more descriptions related to X (this application
is also effective for dialog systems).

In general, the performance of QA systems are evaluated based on
coverage and accuracy. Coverage is the ratio between the number of
questions answered (disregarding their correctness) and the total
number of questions. Accuracy is the ratio between the number of
correct answers and the total number of answers made by the system.

While coverage can be estimated objectively and systematically,
estimating accuracy relies on human subjects (because there is no
absolute description for term X), and thus is expensive.

In view of this problem, we targeted Information Technology Engineers
Examinations\footnote{Japan Information-Technology Engineers
Examination Center. http://www.jitec.jipdec.or.jp/}, which are
biannual (spring and autumn) examinations necessary for candidates to
qualify to be IT engineers in Japan.

Among a number of classes, we focused on the ``Class II'' examination,
which requires fundamental and general knowledge related to
information technology. Approximately half of questions are associated
with IT technical terms.

Since past examinations and answers are open to the public, we can
evaluate the performance of our QA system with minimal cost.

\subsection{Analyzing IT Engineers Examinations}
\label{subsec:analysis}

The Class II examination consists of quadruple-choice questions, among
which technical term questions can be subdivided into two types.

In the first type of question, examinees choose the most appropriate
description for a given technical term, such as ``memory interleave''
and ``router.''

In the second type of question, examinees choose the most appropriate
term for a given question, for which we show examples collected from
the examination in the autumn of 1999 (translated into English by one
of the authors) as follows:
\begin{enumerate}
\item Which data structure is most appropriate for FIFO (First-In
  First-Out)?

  a) binary trees, b) queues, c) stacks, d) heaps
\item Choose the LAN access method in which multiple terminals transmit
  data simultaneously and thus they potentially collide.

  a) ATM, b) CSM/CD, c) FDDI, d) token ring
\end{enumerate}

In the autumn of 1999, out of 80 questions, the number of the first
and second types were 22 and 18, respectively.

\subsection{Implementing a QA system}
\label{subsec:implementation}

For the first type of question, human examinees would search their
knowledge base (i.e., memory) for the description of a given term, and
compare that description with four candidates.  Then they would choose
the candidate that is most similar to the description.

For the second type of question, human examinees would search their
knowledge base for the description of each of four candidate terms.
Then they would choose the candidate term whose description is most
similar to the question description.

The mechanism of our QA system is analogous to the above human
methods.  However, unlike human examinees, our system uses an
encyclopedia generated from the Web as a knowledge base.

In addition, our system selectively uses term descriptions categorized
into domains related to information technology.  In other words, the
description of ``pipeline (transportation pipe)'' is irrelevant or
misleading to answer questions associated with ``pipeline (processing
method).''

To compute the similarity between two descriptions, we used techniques
developed in IR research, in which the similarity between a user query
and each document in a collection is usually quantified based on word
frequencies.  In our case, a question and four possible answers
correspond to query and document collection, respectively.  We used a
probabilistic method~\cite{robertson:sigir-94}, which is one of the
major IR methods.

To sum up, given a question, its type and four choices, our QA system
chooses one of four candidates as the answer, in which the resolution
algorithm varies depending on the question type.

\subsection{Related Work}
\label{subsec:related_work}

Motivated partially by the TREC-8 QA
collection~\cite{voorhees:sigir-2000}, question answering has of late
become one of the major topics within the NLP/IR communities.

In fact, a number of QA systems targeting the TREC QA collection have
recently been
proposed~\cite{harabagiu:coling-2000,moldovan:acl-2000,prager:sigir-2000}.
Those systems are commonly termed ``open-domain'' systems, because
questions expressed in natural language are not necessarily limited to
explicit axes, including {\em who\/}, {\em what\/}, {\em when\/}, {\em
where\/}, {\em how\/} and {\em why}.

However, Moldovan and Harabagiu~\shortcite{moldovan:acl-2000} found
that each of the TREC questions can be recast as either a single axis
or a combination of axes.  They also found that out of the 200 TREC
questions, 64 questions (approximately one third) were associated with
the {\em what\/} axis, for which the Web-based encyclopedia is
expected to improve the quality of answers.

Although Harabagiu~\etal~\shortcite{harabagiu:coling-2000} proposed a
knowledge-based QA system, most existing systems rely on conventional
IR and shallow NLP methods. The use of encyclopedic knowledge for QA
systems, as we demonstrated, needs to be further explored.

\section{Experimentation}
\label{sec:experimentation}

\subsection{Methodology}
\label{subsec:eval_method}

We conducted a number of experiments to investigate the effectiveness
of our methods.

First, we generated an encyclopedia by way of our Web-based method (see
Sections~\ref{sec:system_design} and \ref{sec:organization}), and
evaluated the quality of the encyclopedia itself.

Second, we applied the generated encyclopedia to our QA system (see
Section~\ref{sec:application}), and evaluated its performance.  The
second experiment can be seen as a task-oriented evaluation for our
encyclopedia generation method.

In the first experiment, we collected 96 terms from technical term
questions in the Class II examination (the autumn of 1999). We used as
test inputs those 96 terms and generated an encyclopedia, which was
used in the second experiment.

For all the 96 test terms, Google (see Section~\ref{subsec:retrieval})
retrieved a positive number of pages, and the average number of pages
for one term was 196,503. Since Google practically outputs contents of
the top 1,000 pages, the remaining pages were not used in our
experiments.

In the following two sections, we explain the first and second
experiments, respectively.

\subsection{Evaluating Encyclopedia Generation}
\label{subsec:eval_generation}

For each test term, our method first computed $P(d|c)$ using
Equation~\eq{eq:organization} and discarded domains whose $P(d|c)$ was
below 0.05. Then, for each remaining domain, descriptions with higher
$P(d|c)$ were selected as the final outputs.

We selected the top three (not one) descriptions for each domain,
because reading a couple of descriptions, which are short paragraphs,
is not laborious for human users in real-world usage. As a result, at
least one description was generated for 85 test terms, disregarding
the correctness.  The number of resultant descriptions was 326 (3.8
per term). We analyzed those descriptions from different perspectives.

\begin{figure}[htbp]
  \begin{center}
    \leavevmode
    \psfig{file=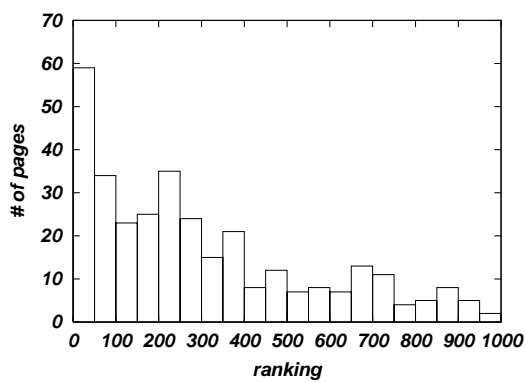,height=2in}
  \end{center}
  \caption{Distribution of rankings for original pages in Google.}
  \label{fig:ranking}
\end{figure}

First, we analyzed the distribution of the Google ranks for the Web
pages from which the top three descriptions were eventually retained.
Figure~\ref{fig:ranking} shows the result, where we have combined the
pages in groups of 50, so that the leftmost bar, for example, denotes
the number of used pages whose original Google ranks ranged from 1 to
50.

Although the first group includes the largest number of pages, other
groups are also related to a relatively large number of pages.  In
other words, our method exploited a number of low ranking pages, which
are not browsed or utilized by most Web users.

Second, we analyzed the distribution of domains assigned to the 326
resultant descriptions.  Figure~\ref{fig:domain_dist} shows the
result, in which, as expected, most descriptions were associated with
the computer domain.

However, the law domain was unexpectedly associated with a relatively
great number of descriptions.  We manually analyzed the resultant
descriptions and found that descriptions for which appropriate domains
are not defined in our domain model, such as sports, tended to be
categorized into the law domain.

\begin{figure}[htbp]
  \begin{center}
    \small
    \begin{tabular}{l} \hline\hline
      computers (200),
      law (41),
      electricity (28), \\
      plants (15),
      medicine (10),
      finance (8), \\
      mathematics (8),
      mechanics (5),
      biotechnology (4), \\
      construction (2),
      ecology (2),
      chemistry (1), \\
      energy (1),
      oceanography (1) \\
      \hline
    \end{tabular}
    \caption{Distribution of domains related to the 326 resultant 
    descriptions.}
    \label{fig:domain_dist}
  \end{center}
\end{figure}

Third, we evaluated the accuracy of our method, that is, the quality
of an encyclopedia our method generated.  For this purpose, each of
the resultant descriptions was judged as to whether or not it is a
correct description for a term in question. Each domain assigned to
descriptions was also judged correct or incorrect.

We analyzed the result on a description-by-description basis, that is,
all the generated descriptions were considered independent of one
another. The ratio of correct descriptions, disregarding the domain
correctness, was 58.0\% (189/326), and the ratio of correct
descriptions categorized into the correct domain was 47.9\% (156/326).

However, since all the test terms are inherently related to the IT
field, we focused solely on descriptions categorized into the computer
domain.  In this case, the ratio of correct descriptions, disregarding
the domain correctness, was 62.0\% (124/200), and the ratio of correct
descriptions categorized into the correct domain was 61.5\% (123/200).

In addition, we analyzed the result on a term-by-term basis, because
reading only a couple of descriptions is not crucial.  In other words,
we evaluated each term (not description), and in the case where at
least one correct description categorized into the correct domain was
generated for a term in question, we judged it correct.  The ratio of
correct terms was 89.4\% (76/85), and in the case where we focused
solely on the computer domain, the ratio was 84.8\% (67/79).

In other words, by reading a couple of descriptions (3.8 descriptions
per term), human users can obtain knowledge of approximately 90\% of
input terms.

Finally, we compared the resultant descriptions with an existing
dictionary. For this purpose, we used the ``Nichigai'' computer
dictionary~\cite{nichigai_compdic:96}, which lists approximately
30,000 Japanese technical terms related to the computer field, and
contains descriptions for 13,588 terms.  In the Nichigai dictionary,
42 out of the 96 test terms were described. Our method, which
generated correct descriptions associated with the computer domain for
67 input terms, enhanced the Nichigai dictionary in terms of quantity.

These results indicate that our method for generating encyclopedias is
of operational quality.

\subsection{Evaluating Question Answering}
\label{subsec:eval_qa}

We used as test inputs 40 questions, which are related to technical
terms collected from the Class II examination in the autumn of 1999.

The objective here is not only to evaluate the performance of our QA
system itself, but also to evaluate the quality of the encyclopedia
generated by our method.

Thus, as performed in the first experiment
(Section~\ref{subsec:eval_generation}), we used the Nichigai computer
dictionary as a baseline encyclopedia. We compared the following three
different resources as a knowledge base:
\begin{itemize}
\item the Nichigai dictionary (``Nichigai''),
\item the descriptions generated in the first experiment (``Web''),
\item combination of both resources (``Nichigai + Web'').
\end{itemize}

Table~\ref{tab:eval_qa} shows the result of our comparative
experiment, in which ``C'' and ``A'' denote coverage and accuracy,
respectively, for variations of our QA system.

Since all the questions we used are quadruple-choice, in case the
system cannot answer the question, random choice can be performed to
improve the coverage to 100\%.  Thus, for each knowledge resource we
compared cases without/with random choice, which are denoted ``w/o
Random'' and ``w/ Random'' in Table~\ref{tab:eval_qa}, respectively.

\begin{table}[htbp]
  \begin{center}
    \caption{Coverage and accuracy (\%) for different question
    answering methods.}
    \medskip
    \leavevmode
    \small
    \begin{tabular}{lcccc} \hline\hline
      & \multicolumn{2}{c}{w/o Random} & \multicolumn{2}{c}{w/
      Random} \\
      \multicolumn{1}{c}{Resource} &
      C &
      A &
      C &
      A \\ \hline
      Nichigai & 50.0 & 65.0 & 100 & 45.0 \\
      Web & 92.5 & 48.6 & 100 & 46.9 \\
      Nichigai + Web & 95.0 & 63.2 & 100 & 61.3 \\
      \hline
    \end{tabular}
    \label{tab:eval_qa}
  \end{center}
\end{table}

In the case where random choice was not performed, the Web-based
encyclopedia noticeably improved the coverage for the Nichigai
dictionary, but decreased the accuracy.
However, by combining both resources, the accuracy was noticeably
improved, and the coverage was comparable with that for the Nichigai
dictionary.

On the other hand, in the case where random choice was performed, the
Nichigai dictionary and the Web-based encyclopedia were comparable in
terms of both the coverage and accuracy.  Additionally, by combining
both resources, the accuracy was further improved.

We also investigated the performance of our QA system where
descriptions related to the computer domain are solely used. However,
coverage/accuracy did not significantly change, because as shown in
Figure~\ref{fig:domain_dist}, most of the descriptions were inherently
related to the computer domain.

\section{Conclusion}
\label{sec:conclusion}

The World Wide Web has been an unprecedentedly enormous information
source, from which a number of language processing methods have been
explored to extract, retrieve and discover various types of
information.

In this paper, we aimed at generating encyclopedic knowledge, which is
valuable for many applications including human usage and natural
language understanding.
For this purpose, we reformalized an existing Web-based extraction
method, and proposed a new statistical organization model to improve
the quality of extracted data.

Given a term for which encyclopedic knowledge (i.e., descriptions) is
to be generated, our method sequentially performs a) retrieval of Web
pages containing the term, b) extraction of page fragments describing
the term, and c) organizing extracted descriptions based on domains
(and consequently word senses).

In addition, we proposed a question answering system, which answers
interrogative questions associated with {\it what\/}, by using a
Web-based encyclopedia as a knowledge base.  For the purpose of
evaluation, we used as test inputs technical terms collected from the
Class II IT engineers examination, and found that the encyclopedia
generated through our method was of operational quality and quantity.

We also used test questions from the Class II examination, and
evaluated the Web-based encyclopedia in terms of question
answering. We found that our Web-based encyclopedia improved the
system coverage obtained solely with an existing dictionary. In
addition, when we used both resources, the performance was further
improved.

Future work would include generating information associated with more
complex interrogations, such as ones related to {\it how\/} and {\it
why\/}, so as to enhance Web-based natural language understanding.

\section*{Acknowledgments}

The authors would like to thank NOVA, Inc. for their support with the
Nova dictionary and Katunobu Itou (The National Institute of Advanced
Industrial Science and Technology, Japan) for his insightful comments
on this paper.

\bibliographystyle{acl}

\end{document}